# Rethinking Pretraining for Specialized Design Data: Evidence from the JONES-19 Cultural Design Dataset

Alexandros Haridis and Charles Zhou

Harvard University, Cambridge, MA 02139, USA
haridis@seas.harvard.edu

**Abstract.** Design and architectural archives encode expert human knowledge in graphical formats, providing a critical testbed for design-inspired Machine Learning (ML) challenges absent with typical computer vision benchmarks. Building on JONES-19, a small-size image dataset based on *The Grammar of Ornament* (London, 1857), we evaluate the discriminative performance of Convolutional Neural Networks (CNNs) in two model training strategies: (a) ImageNet pretraining for domain-general "visual common sense," and (b) learning from scratch on the design data in JONES-19. We find that while domain-general priors improve discriminative performance, learning from scratch augmented with repeated local sampling (multi-crop) effectively recovers these gains. For highly structured design data, local design-driven representations provide sufficient foundation for learning, challenging a reliance on massive general-purpose pretraining. These findings suggest that in specialized design domains, careful curation of smaller high-quality datasets that capture empirical and formal design principles may prove more effective and informative on the nature of a particular design domain than prioritizing large-scale data collection.

## 1 Introduction

Design and architectural archives represent repositories of expert design knowledge encoded in graphical formats—drawings, diagrams, patterns. When digitized, these design collections offer new opportunities at the intersection of Machine Learning (ML) and design scholarship. They provide a critical testbed for design-inspired ML challenges that may not arise with typical computer vision benchmarks while opening new avenues for understanding human intelligence in creative domains.

This paper examines how domain-general visual knowledge—acquired through ImageNet pretraining—influences the discriminative performance of Convolutional

*This is a preprint of a paper published in the Proceedings of the 2026 Design Computing and Cognition Conference (J.S. Gero and T. Shealy (eds)) held in Paris, France, July 6–10, 2026. Please visit the publisher's website for the final version of record.*

Neural Networks (CNNs) on JONES-19, a small-size dataset of ornament designs based on *The Grammar of Ornament* (1857) by Owen Jones [1, 2, 17]. To what extend is the "visual common sense" derived from millions of natural images necessary to capture the visual representations inherent in a specialized design collection of abstract ornament designs? We investigate this tension through the lens of JONES-19 by comparing two training strategies: (a) leveraging ImageNet pretraining, and (b) learning from scratch solely on JONES-19. We devise controlled experiments with ResNet18 and ResNet50 to evaluate four conditions: with and without ImageNet-1K pretraining, and with and without multi-crop augmentation.

Our central question is: To what extent does ImageNet pretraining improve classification accuracy on specialized ornament data, and how does its impact compare to performance gains from data augmentation alone? While domain-general priors improve performance, we find that learning from scratch augmented with multi-crop sampling effectively recovers these gains. This suggests that for highly structured design data, local design-driven representations provide a foundation for sufficient learning, challenging assumptions about the necessity of large-scale pretraining.

These findings have practical implications for specialized domains: careful curation of smaller, high-quality datasets reflecting formal design principles may prove more effective than prioritizing dataset scale. Building on the original JONES-19 publication's baseline benchmarks, this work advances our understanding of how models acquire and transfer visual knowledge across domains and aims to motivate joint investigation by the design and ML communities into questions leading toward more human-aligned intelligence inspired by design and architectural principles.

## 2 Background

### 2.1 The JONES-19 Dataset

The authors of the JONES-19 dataset paper [1], argue that design archives are sources of expert design knowledge encoded in high-quality graphical formats—drawings, pictures, diagrams, etc. *The Grammar of Ornament*, and similar collections, provide a testbed for design-inspired learning challenges that don't necessarily arise with typical ML benchmarks. JONES-19 is defined by several distinct characteristics:

- A small, class-imbalanced sample size of 1,901 high-resolution images, reflecting the natural scarcity of specialized archival data (Figure 1).
- The dataset exclusively documents ornament designs from nineteen cultures—a specialized class of human-designed artifacts that differ fundamentally from the "objects-in-context" found in natural scene datasets (e.g., "a cat on a sofa").
- Designs are defined by fine artistic details, including diverse line and color patterns, geometric motifs, and cultural themes. These are fundamental human-designed descriptors of a culture rather than "image noise."

- The ornaments have non-uniform sizes and shapes, involving non-rectangular boundaries (Figure 2). By comparison, the ImageNet-1k dataset [3, 4], has more than a million images with canonical rectangular boundaries.

Figure 1 (a, b) shows a subset of the JONES-19 dataset with a class distribution histogram. The Roman class, for example, with its intricate hand-drawn rock carvings, comes with just twelve image samples, illustrating a typical case of design data produced purely by manual human effort in surveying. Such “long-tail” cases, challenge models to learn from extreme sparsity.

Previous work [1] demonstrates that CNNs learn feature representations of the ornament designs in JONES-19 that effectively partition the data into their corresponding classes. We visualize this learned embedding space in Figure 3, providing qualitative evidence of the discriminative capabilities of these models. The summary statistics in Figure 4 show that while a large subset of images has standard rectangular or square dimensions across nearly all cultures, a substantial portion of ornaments resists this categorization. Many ornament designs are portrayed with non-canonical boundary shapes and irregular scales, reflecting their presentation on the original plates (e.g., Figure 2). This structural variety directly impacts standard augmentation techniques during the training phase—such as fixed-ratio resizing and cropping.

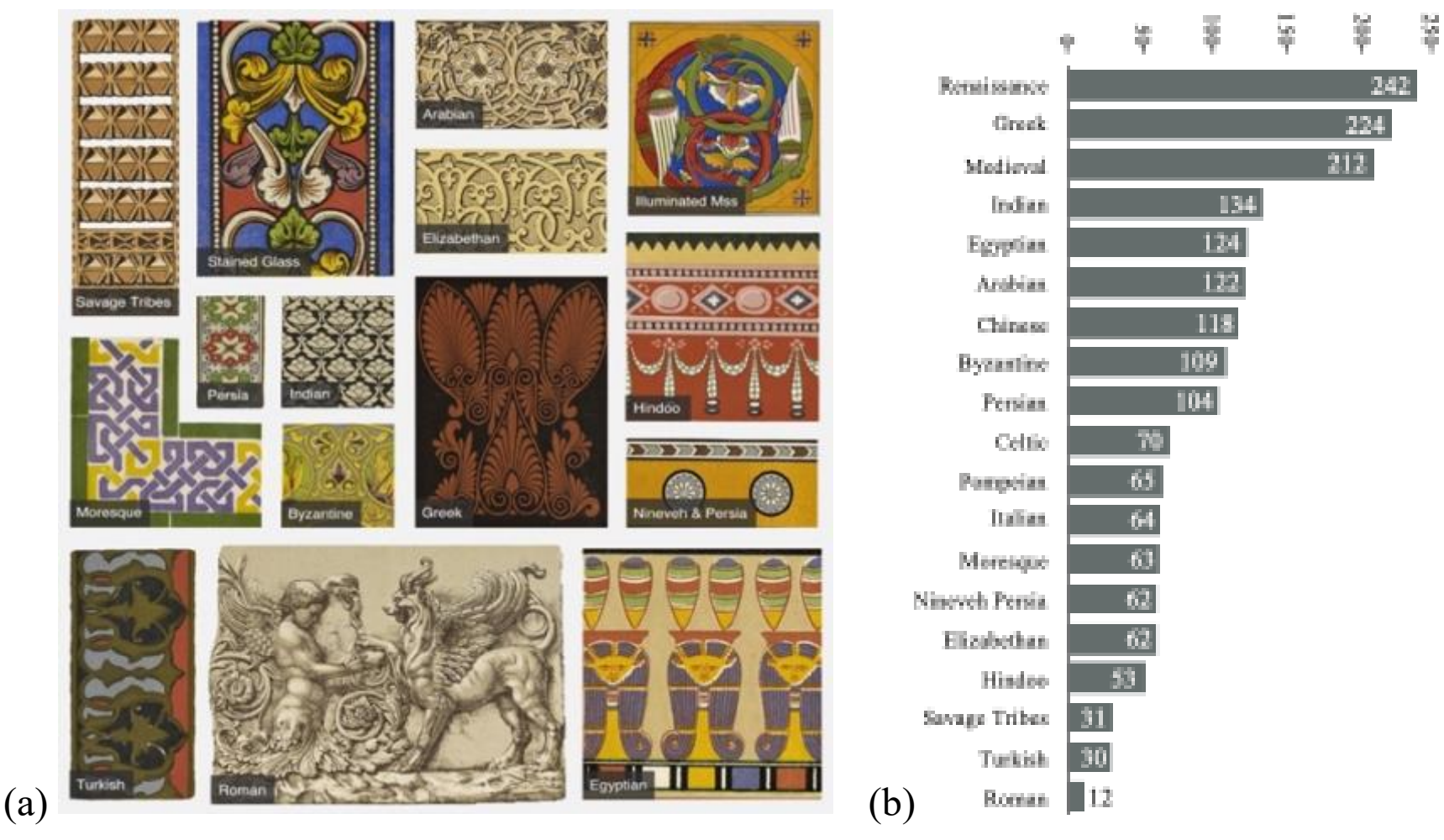

**Fig. 1.** (a) A subset of the JONES-19 image dataset [2]. (b) Class distribution across 19 cultures.

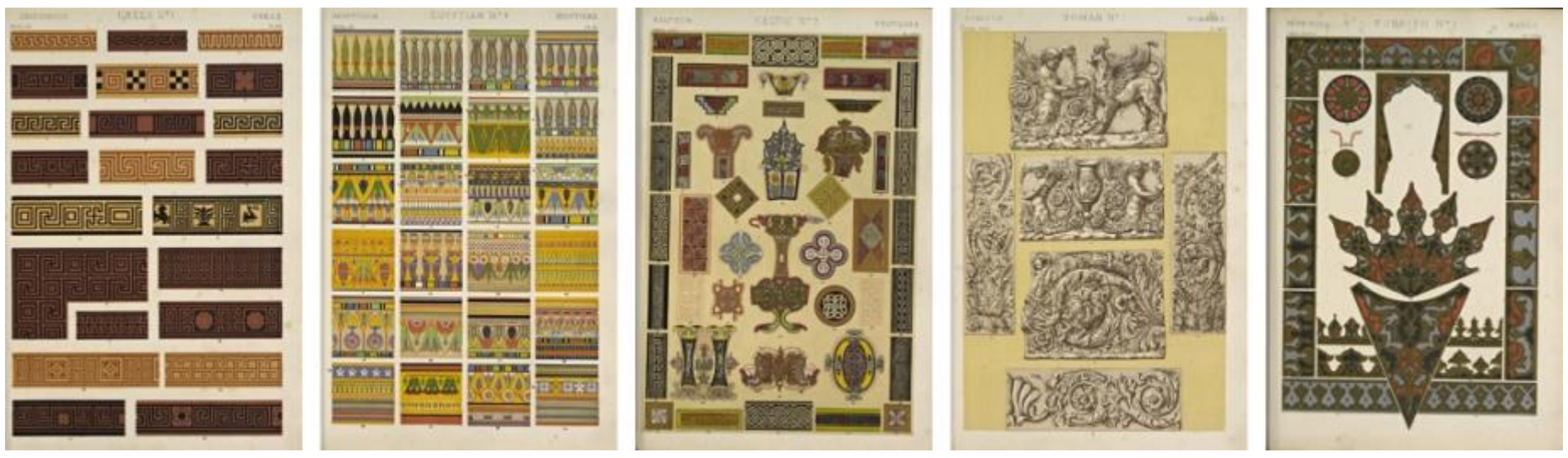

**Fig. 2.** Sample pages of the Open Access digital version of *The Grammar of Ornament*; see [1].

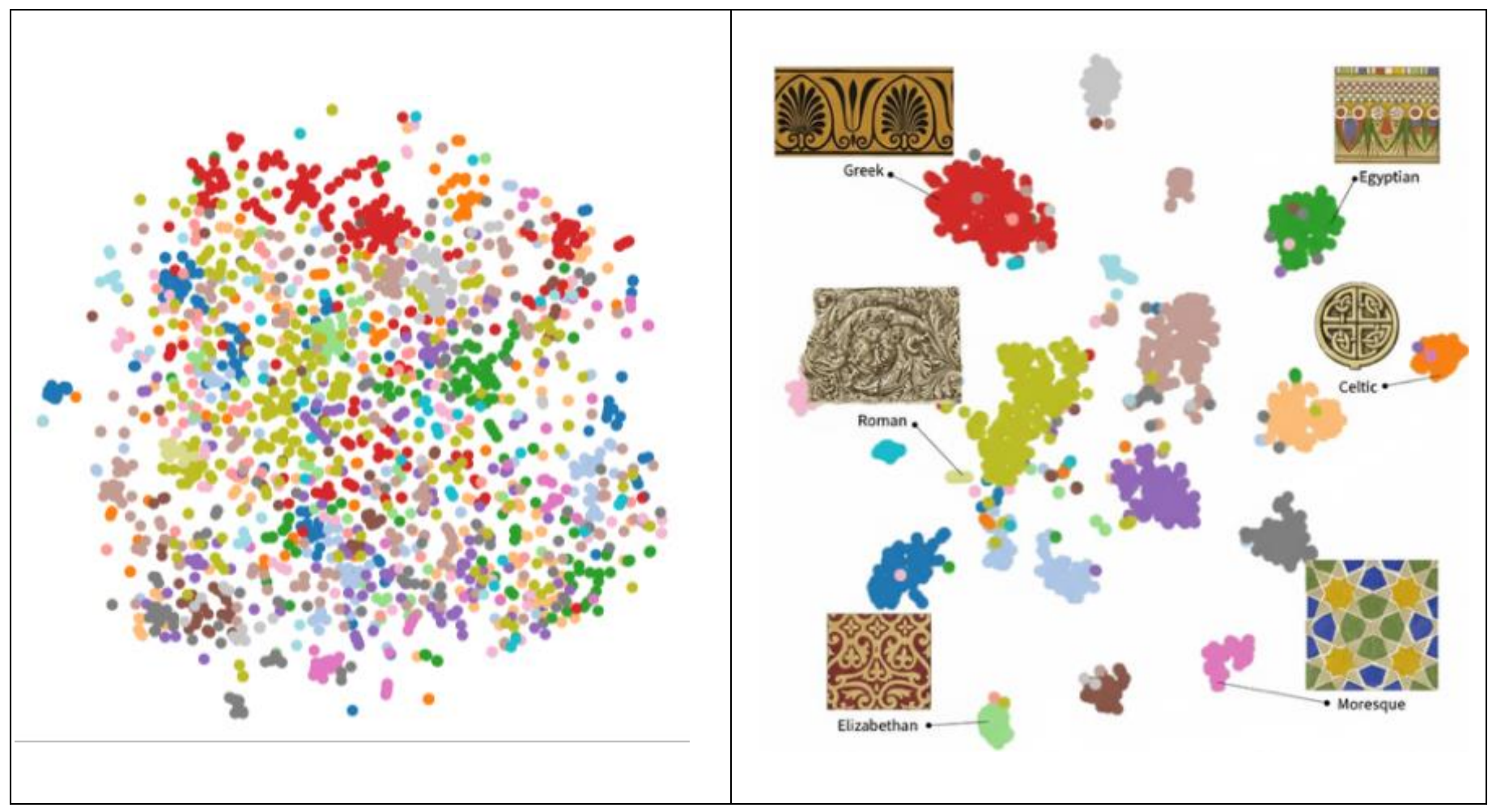

**Fig. 3.** Visualization of the learned embedding space using t-SNE (p=30) before (left) and after (right) finetuning. The 2D projections are a qualitative assessment of how well the classes in JONES-19 are separated in latent space by the ResNet-18 model reported in [1].

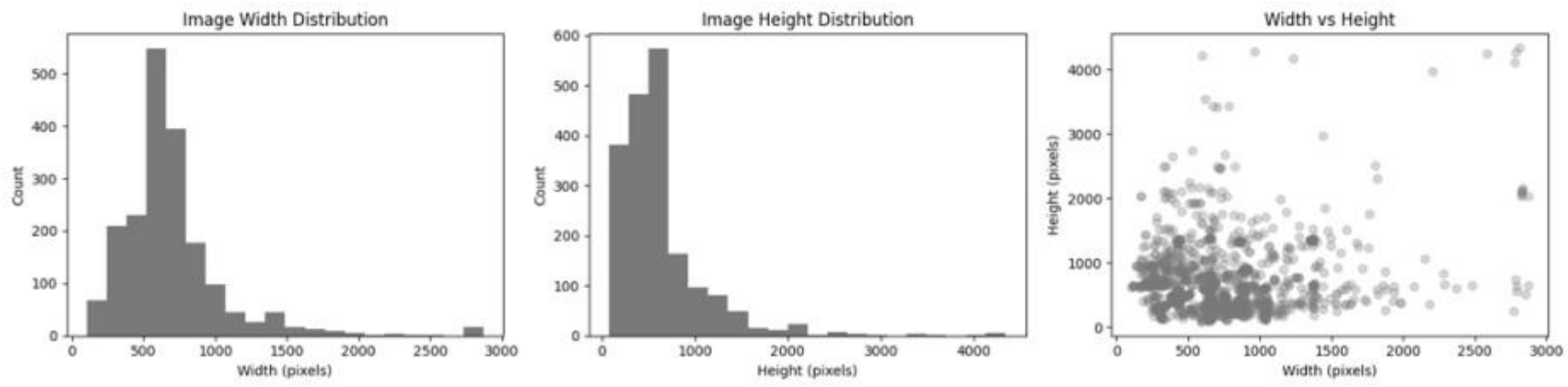

**Fig. 4.** Fig. 5. Overview of image width and height distributions in JONES-19. The measures represent the size of the outermost rectangle containing an ornament design.

## 2.2 From Objects-In-Context to Human-Made Artifacts

Modern CV and ML research focuses on vision models that exhibit general-purpose visual capabilities, such as object recognition, localization, and image classification.

To achieve this, a model has learned domain-general visual representations after being trained on a substantially large and diverse image dataset. The ImageNet database [3, 4], an important testbed for constructing general-purpose CV, uses the hierarchical structure of WordNet to populate tens of thousands of concepts with 500-1000 images on average illustrating each concept. The concepts that the database illustrates capture a "visual common sense": objects as we find them in the real world, such as "mammal," "bird," "vehicle," "furniture," etc. In this sense, general-purpose databases such as ImageNet enable vision models to tell with human-level accuracy "what" an object is, in the way it is seen in its respective real-world context [7, 8, 15].

The images in JONES-19, on the other hand, are characterized by structured arrangements of line forms and color, often created to imitate Nature's forms – flowers, rocks, trees, etc.—not as a reality, but as an ideal (abstract) representation. Ornament designs are human-made geometrical constructions, often provided in definite proportions. Given that these images are fundamentally different from the "objects-in-context" found in natural scene datasets, we pose the following question that distinguishes between acquiring a visual common sense and acquiring a so called "specialized design perception" for interpreting human-made artifacts: Are the image priors acquired from extensive exposure to domain-general visual representations necessary for learning meaningful visual representations specific to a specialized design collection? How can we teach a vision model to discern "how" a particular design motif, or its style, is constructed—its composition or grammar? We investigate these questions in Section 3 by studying the effect of large-scale ImageNet pretraining on JONES-19. Our objective is to quantify how much classification performance improves due to domain-general pretraining, and to what extent similar gains can be recovered through data augmentation alone on a randomly initialized model.

### 2.3 Pretraining, Data Augmentation, and Small-Size Dataset Learning

Learning visual representations under constrained dataset sizes is a significant problem in computer vision. Two techniques are typically used: (i) transfer learning from a large-scale pretrained model and (ii) regularization through data augmentation. Below we review evidence on the benefits and limitations of each, with an emphasis on small, imbalanced, and domain-specific datasets.

**Pretraining versus training-from-scratch.** CNNs pretrained on large natural image datasets demonstrate strong transferability across downstream tasks [1]. Early work shows that "lower" and mid-level convolutional features—edges, textures, and simple shapes—are highly transferable across domains, while "higher" layers encode task-specific features [6, 12]. Reusing lower-level features while fine-tuning for a different downstream task is the dominant method of transfer learning, and models initialized from pretrained weights consistently outperform random initializations when labeled data is scarce [4]. Kornblith et al. [7] further demonstrate that stronger pretrained models tend to transfer better even across substantially different domains—a finding we engage with directly by experimenting with ResNet18 and ResNet50.

A recurring finding is that the benefit of pretraining is inversely related to dataset size. In contrast, He et al. [8] find that training-from-scratch can match pretrained performance when sufficient labeled data are available—a condition rarely met in specialized data-scarce domains such as medical imaging, cultural heritage, architecture, visual arts, and art-historical classification (e.g., manuscripts, paintings), where annotation requires domain expertise and data acquisition is inherently limited, often involving ethical issues on copyright and authorship.

**Data Augmentation.** Data augmentation increases the number and diversity of training samples without introducing new semantic labels. Its purpose is to enforce robustness by exposing a model to transformed views of the same image [5]. Empirical results show that augmentation yields larger relative gains on smaller datasets [6]. Geometric transformations, such as cropping (i.e., local sampling), horizontal flipping, and small rotations, are particularly effective for image classification. Cropping strategies can be interpreted as approximating training on a larger dataset, and prior work shows they can close a substantial portion of the performance gap between pretrained and non-pretrained models [6]. While learned augmentation strategies such as *AutoAugment* can outperform manually designed pipelines [10, 11], they are computationally expensive and often require dataset-specific policy search.

A central question in small-data learning is whether aggressive augmentation can substitute for large-scale pretraining. Pretraining supplies strong representational priors; augmentation enforces task-relevant invariances. Evidence suggests diminishing returns when combining very strong pretraining with heavy augmentation, but the combination generally yields the best performance in limited-data settings [6, 7].

**Imbalanced Data.** Small-size datasets that originate from design collections such as JONES-19 are often imbalanced, complicating both training and evaluation. Standard accuracy metrics can be misleading because majority classes dominate optimization [13]. Here, we adopt stratified cross-validation, class-weighted loss functions, and balanced metrics to counteract these effects. Because JONES-19 exclusively contains abstract ornament patterns—geometric human-made motifs, color and texture, fine stylistic details—it presents challenges beyond dataset size and class imbalance. We hypothesize that while low-level features remain transferable in this setting, higher-level semantic features may be less relevant, making pretraining a strong but imperfect transfer and justifying our use of domain-appropriate augmentation strategies.

## 3 Evaluating Vision Models: Methodology and Results

### 3.1 The Learning Task

We evaluate the discriminative performance of CNNs under two model training strategies: (a) ImageNet pretraining, and (b) training-from-scratch on JONES-19 with randomly initialization. In both settings, we apply identical data augmentation to isolate the effects of pretraining and augmentation in a small, domain-specific dataset. By varying pretraining (present vs. absent) and augmentation strength (single- vs. multi-crop) while holding architecture and training constant, we evaluate whether extensive augmentation can compensate for the absence of pretraining and how these factors interact to influence classification performance. Supplementary material is available in an external repository (see Acknowledgments).

### 3.2 Experimental Setup

In all experiments, we use the same residual network architectures as in the original JONES-19 paper [1]: ResNet-18 and Res-Net-50 [9]. Pretrained models are initialized using weights learned on ImageNet-1K [3]. In training-from-scratch experiments, the weights are initialized using Kaiming-normal initialization [15]. All models are trained end-to-end to predict one of the nineteen ornament classes.

**Data Preprocessing and Augmentation.** All input images are resized to 256 pixels on their shorter side and cropped to 224x224 resolution. Training images are randomly cropped, while test images are center-cropped. Pretrained models use input images with ImageNet channel normalization and standard deviation; models trained from scratch do not.

While validation images undergo deterministic preprocessing only (i.e., square-shape center-crop and ImageNet normalization), training images undergo stochastic geometric augmentations: random horizontal flips in 50% of the cases and random rotation of up to ±5°. JONES-19 contains images significantly larger than the input resolutions. This allows us to take advantage of each image to yield multiple unique training examples per ornament through cropping. Thus, in addition to the above geometric augmentations, we distinguish between two augmentation strategies:

- Single-crop: each image contributes one stochastic view per epoch.
- Multi-crop: each image contributes seven (7) independently augmented views per epoch.

Following [1], who estimated that the average image size of 680x555 in JONES-19 permits approximately 7 crops per image, we implement multi-crop training via a dataset wrapper that samples multiple transformed instances per image, increasing the effective training set from ~1900 to ~13300 samples per fold while preserving label distributions. Validation always uses a single deterministic crop to ensure accuracy reflects performance on the ornament designs in JONES-19.

**Training Procedure.** Given the high class imbalance, we adopt a stratified 10-fold cross-validation, ensuring that each class appears in every validation split. Models are trained for 40 epochs using the Adam optimizer [16] with weighted cross-entropy loss, where class weights are inversely proportional to class frequency. Classification performance is measured via balanced accuracy—the average of per-class accuracies.

and each experiment is repeated 10 times. We repeat each experiment 10 times and report the average accuracy along with its standard deviation.

### 3.3 Results

Tables 1 and 2, summarize the results of our experiments. The tables report the balanced accuracies for our two baseline models, each one under two distinct training strategies. ImageNet pretraining yields large performance gains under single-crop augmentation, with accuracies between 77% (± 4.03) and 79% (± 2.85). This shows a significant improvement over random guessing, which would yield approximately 15% accuracy for a nineteen-class learning problem. Compared to single-crop augmentation with training-from-scratch, ImageNet pretraining improves balanced accuracy by approximately 18-22 percent points across both architectures.

Most notably, training-from-scratch with multi-crop sampling approaches the accuracy of pretrained models under single-crop augmentation, particularly for Resnet-18. This indicates that repeated image sampling can partially substitute for large-scale pretraining in the case of JONES-19, which represents a data-scarce domain with highly specialized and structured design content. Pretrained models still outperform under multi-crop augmentation, although the gains are less pronounced compared to the single-crop setting (for example, the pretrained ResNet-18 improves over no-pretraining by just 3%). The additional 3–7% accuracy gain observed with ImageNet in a multi-crop scenario indicates that domain-general visual representations remain beneficial even when training sample diversity is increased. A more definitive account of these effects would require investigating whether ImageNet priors are suboptimal or introduce biases when applied to abstract design patterns, which we leave for future work.

**Table 1.** ResNet-18 (224×224) mean balanced accuracy across different training and augmentation strategies.

| Augmentation | No Pretraining | ImageNet Pretraining |
|---|---|---|
| Single-crop | 59.07 ± 6.71 | 77.34 ± 4.03 |
| Multi-crop | 75.31 ± 5.96 | 78.76 ± 6.12 |

**Table 2.** ResNet50 (224×224) mean balanced accuracy across different training and augmentation strategies.

| Augmentation | No Pretraining | ImageNet Pretraining |
|---|---|---|
| Single-crop | 58.03 ± 9.18 | 79.61 ± 2.85 |
| Multi-crop | 77.91 ± 4.25 | 84.11 ± 3.95 |

ResNet-50 consistently outperforms ResNet-18 across all scenarios, with the largest gains observed when combined with both pretraining and multi-crop augmentation. This result indicates that increased model capacity (e.g., deeper architectures and number of parameters) better exploits both general-domain visual priors acquired through pretraining and augmented training diversity. This holds even for ornament designs, which exhibit intricate details such as variations in line thickness, relief, color, texture, and local symmetry.

The confusion matrices corresponding to ResNet-50's performance in the ImageNet pretraining and training-from-scratch scenarios are printed in Figure 5. Using these matrices, we can compare the pretraining and no-pretraining scenarios in detail by directly observing model performance on individual classes.

In particular, multi-crop augmentation improves recall for several classes but does not fully resolve fine-grained cross-cultural similarities that lead to confusions. The pretrained ResNet-50 combined with multi-crop augmentation achieves diagonal dominance in 15 out of 19 classes, with ResNet-18 performing the same or better in the remaining 4 classes. Even with an aggressive augmentation strategy, ResNet-50 still achieves less than 65% accuracy in the Persian, Chinese, and Indian ornament with approximately 100 image samples each. By comparison, the same model achieves 82% and 95% accuracy for the Medieval and Greek cultures, respectively, each one with approximately 200 image samples.

(a) ResNet50, 224 resolution, no-pretraining, multicrop
Acc: 77.91% ±4.25%

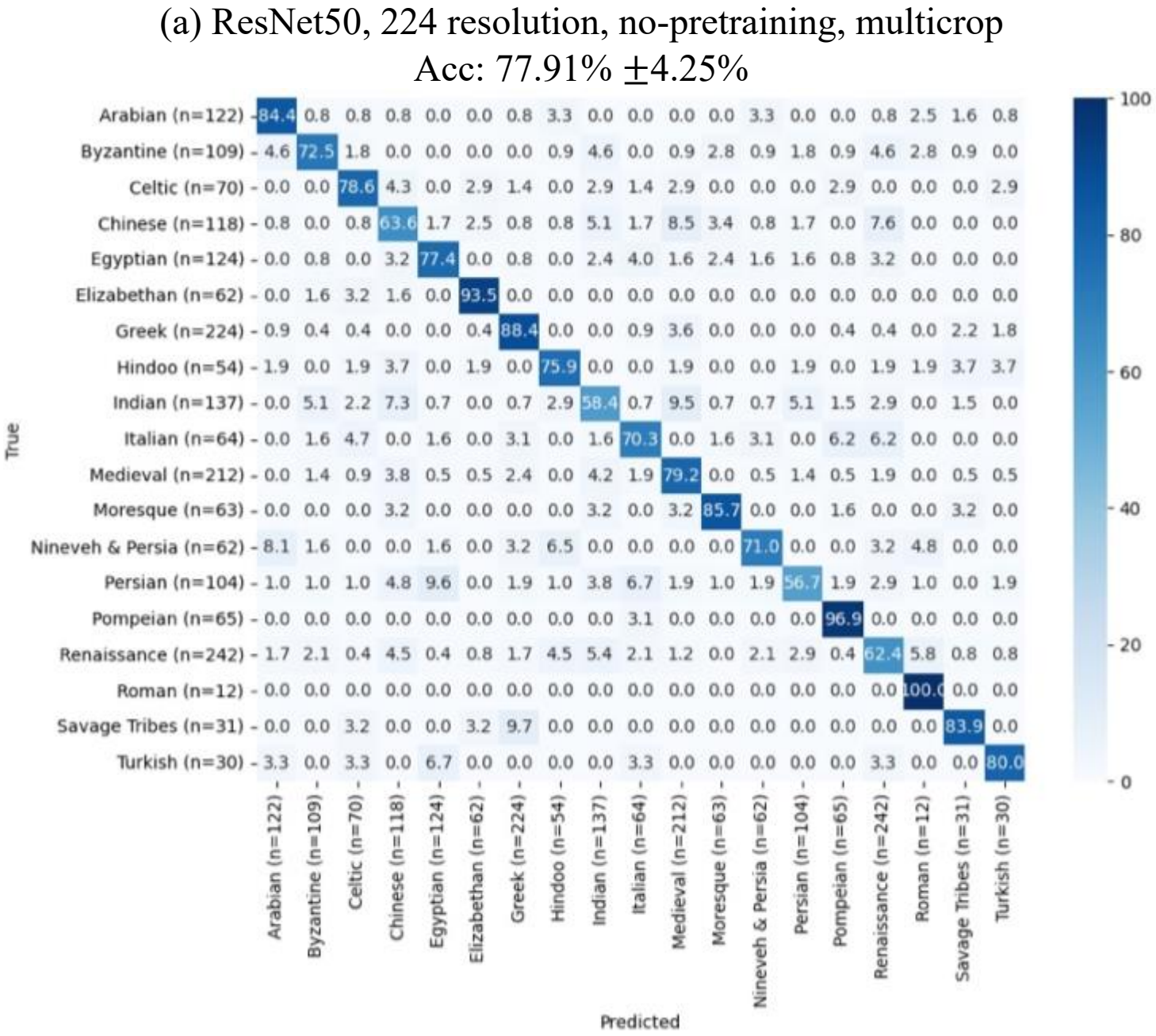

(b) ResNet50, 224 resolution, ImageNet-pretraining, multicrop
Acc: 84.11% ±3.95%

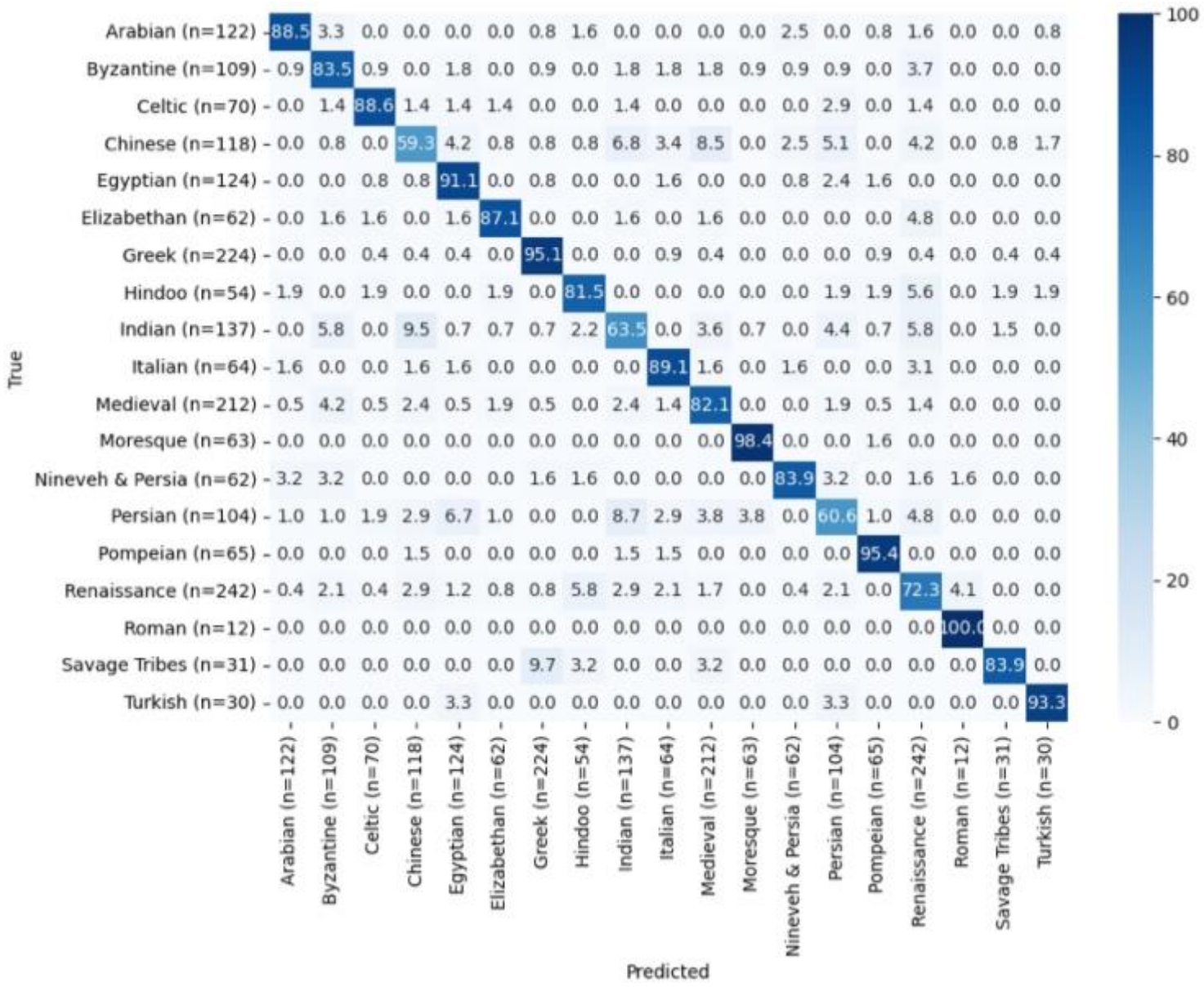

**Fig. 6.** Confusion matrices for two variants of the ResNet-50 baseline classifier on JONES-19: (a) training-from-scratch, and (b) ImageNet pretraining. The matrices show the performance of the best model across all runs. We provide additional results in the supplementary material.

Both models achieve perfect score in the Roman class with just 12 image samples, while ResNet-50 improves the accuracy on the Turkish ornament, the second lowest sample class, by 13 percent points. These error results can't be attributed to data scarcity alone (i.e., number of samples in a class). Many other characteristics of ornament designs play a major contributing factor to model confusion. Examples are the internal structural coherence of intra-class samples (e.g., the consistency of the Roman class), cross-cultural visual similarities (e.g., that the Assyrian ornament was a borrowed style from the Egyptian [17, p.89]), as well as the method of their depiction on the pages of the original book (i.e., the lines and colors used by Owen Jones).

# 4 Discussion

## 4.1 Design-driven Representations

While ImageNet pretraining provides clear benefits under standard training conditions, these gains can largely be recovered through intensive local sampling when training-from-scratch on a domain-specific dataset. Our main result shows that training-from-scratch with multi-crop augmentation approaches the performance of ImageNet-pretrained models under single-crop conditions, particularly for ResNet-18.

The effectiveness of multi-crop sampling suggests that ornament designs contain rich local structure that models can exploit without relying on the broad visual vocabulary previously learned from natural images. This has important implications for design and architectural collections, where large-scale labeled datasets are often impractical. Rather than prioritizing scale, we propose emphasizing the construction of smaller, high-quality datasets that capture the formal and empirical principles of a domain, combined with intensive sampling of part-structures. Concretely, this involves: (1) careful curation to capture formal structure and empirical principles; (2) augmentation strategies that respect domain constraints while increasing sample diversity; and (3) critical awareness of how dataset construction shapes model behavior.

### 4.2 General Visual Priors

ImageNet pretraining provides consistent performance gains across all experimental conditions. This indicates that domain-general visual features—edges, contours, and spatial relations—remain valuable even for abstract design data. The source of this advantage remains unclear. Pretraining may offer transferable representations that accelerate learning, analogous to how a human with a developed visual perception learns to discern domain-specific artifacts. Alternatively, it may introduce biases that align fortuitously with ornament patterns or simply provide more effective weight initialization. Comparisons across dataset sources—such as synthetic 2D/3D data, digitized artifacts, or alternative design archives—could clarify whether ImageNet's visual vocabulary is inherently necessary for acquiring new visual knowledge.

### 4.3 Beyond Data Scarcity

Performance differences cannot be explained by sample size alone. Some cultural styles exhibit strong internal consistency, allowing models to learn discriminative features from limited data, while others display greater variability and require larger datasets. Historical design traditions frequently borrow across cultures further complicating classification; Owen Jones describes these influences extensively [17]. When styles share visual characteristics, whether through historical influence or convergent geometric principles, models struggle to distinguish them regardless of dataset size. Modeling these cross-cultural visual similarities remains an open problem. The JONES-19 dataset itself derives from a single historical source, and its rendering techniques, color choices, and cropping introduce systematic biases that may amplify or obscure distinctions. This underscores a broader challenge in algorithmic learning: design collections encode curatorial and representational decisions as much as they represent underlying traditions. Models ultimately learn these representations, which may differ subtly but significantly from the underlying cultural reality.

Several limitations frame our findings. JONES-19 is a small, single-source dataset that may not generalize to other design collections. Our augmentation strategy is relatively standard; more specialized augmentations may yield further gains in learning. We also evaluate only ResNet architectures; alternative models, such as transformers, may behave differently.

Future work should test these results across diverse design collections (e.g., architectural drawings, typography) and alternative data generation methods, including synthetic 2D/3D pipelines (e.g., [18]). Comparing ImageNet with design-specific pretraining would further clarify the role of general visual priors. Additionally, interpretability methods that link model decisions to design-theoretic concepts — such as symmetry, part structure, and grammatical rules—could strengthen connections between ML and design scholarship more productively.

From a computational perspective, purely data-driven learning, even when augmented, may be inherently limited in capturing the rule-based, generative structure of design languages. A promising direction is to complement algorithmic learning with explicit rule-based frameworks, such as shape grammars, which encode how designs are generated and composed—knowledge that remains implicit and difficult to extract from standard off-the-shelf neural network models.

## Acknowledgments

We thank Professor Stuart Shieber for providing computing resources for this project, and Helen He for helpful discussions related to this work. Readers may find the Supplementary Material for this publication in the following repository:
https://github.com/alexHaridis/dcc-2026-paris-supplementary-material